\documentclass{article}
\usepackage[T1]{fontenc}
\usepackage{preprint,times}
\usepackage{amsmath,amssymb,booktabs,graphicx,array,tabularx}
\usepackage{hyperref,url,xspace}
\usepackage{placeins}
\hypersetup{hidelinks}
\newcommand{\mname}{\texttt{ViSTA}\xspace}
\newcommand{\R}{\mathbb{R}}
\newcommand{\LN}{\operatorname{LN}}
\newcommand{\MHA}{\operatorname{MHA}}
\newcommand{\RMS}{\operatorname{RMS}}
\newcommand{\FFN}{\operatorname{FFN}}
\title{\mname{}: A Simple Bridge Extends Visual\\Alignment to Clinical Time-Series\\Understanding in Multimodal LLMs}
\author{%
Junyi Gao\textsuperscript{1,3,*}, Yu Shi\textsuperscript{2,*}, Pingzhao Hu\textsuperscript{4,\dag}, Ewen M Harrison\textsuperscript{1,\dag}\\[0.4em]
\normalfont\textsuperscript{1}Centre for Medical Informatics, University of Edinburgh\\
\normalfont\textsuperscript{2}Biostatistics Division, Dalla Lana School of Public Health, University of Toronto\\
\normalfont\textsuperscript{3}Health Data Research UK\\
\normalfont\textsuperscript{4}Department of Biochemistry, Western University\\
\normalfont\textsuperscript{*}Equal contribution\\
\normalfont\textsuperscript{\dag}Senior author\\
}
\begin{document}
\maketitle
\begin{abstract}
Clinical prediction models estimate risk from patient measurements, while large language models support medical text understanding and question answering. Yet their language capabilities do not ensure accurate prediction from structured, high-dimensional clinical time series. Improving this ability would connect risk estimation with flexible questions about a patient's evolving condition. We introduce \mname{}, a compact adapter that incorporates irregular numerical measurements into a pretrained vision-language model's chart representations. It learns corrections to visual tokens while leaving all pretrained parameters unchanged. On MIMIC-IV, \mname{} has the highest mean scores among the compared adaptations on all four metrics for acute kidney injury and mortality prediction across models with 2-9 billion parameters. With 0.516 million trainable parameters, the 2-billion-parameter model reaches an area under the ROC curve of 0.7376 for acute kidney injury, compared with GPT-5.6 Sol's 0.7380 with text input and high reasoning effort. Training for temporal question answering yields 69.27\% accuracy at 4 billion parameters with over 90\% fewer trainable parameters than low-rank adaptation using charts or numerical text, at a 2.82-4.88 percentage-point accuracy gap. \mname{} extends pretrained language models to numerical prediction and temporal questions.
\end{abstract}

\section{Introduction}
Clinical artificial intelligence supports outcome prediction and patient-record interpretation. Statistical and deep learning models use patient measurements to estimate mortality risk, anticipate deterioration, and predict length of stay \citep{clinical_timeseries,gao2020dr,ma2020concare}. Large language models (LLMs) support clinical summarization and medical question answering \citep{clinical_summarization,medical_llms}, offering flexible queries that complement predefined prediction targets.

Extending this flexibility to numerical records remains difficult. Vital signs, laboratory tests, and interventions form high-dimensional time series with irregular sampling, missing values, and dependencies across variables. Their numerical and temporal relationships differ from information in clinical narratives. Recent work finds that even state-of-the-art closed-source LLMs remain behind conventional models on non-generative clinical prediction from structured records, with larger gaps for smaller open-source LLMs \citep{clinicrealm}. Improving this capability would let a language-based system use the same measurements to estimate risk and answer questions about threshold crossings or changes across trajectories, which conventional prediction models cannot do~\citep{zhu2026augmenting}. This motivates adaptation that strengthens numerical prediction while retaining a language interface for patient-record questions.

Existing approaches differ in how they present numerical records and which components they adapt. Text serialization exposes exact measurements but spreads temporal relationships across long sequences of names, timestamps, and values. Learned numerical encoders provide compact representations, but these must be aligned with the language model. Charts offer an established visual-language pathway, yet access to precise measurements depends on what survives rendering and visual encoding \citep{images_signals,mllm4ts}. Fine-tuning and low-rank adaptation (LoRA) require adjustment to language-model weights \citep{lora,forgetting}. These approaches leave an opportunity to combine direct numerical access with the visual-language alignment already learned during pretraining. Can a small adapter enrich the model's existing chart representations with precise measurements while keeping its pretrained components and language output interface unchanged?

\begin{figure}[t]
\centering
\includegraphics[width=\linewidth]{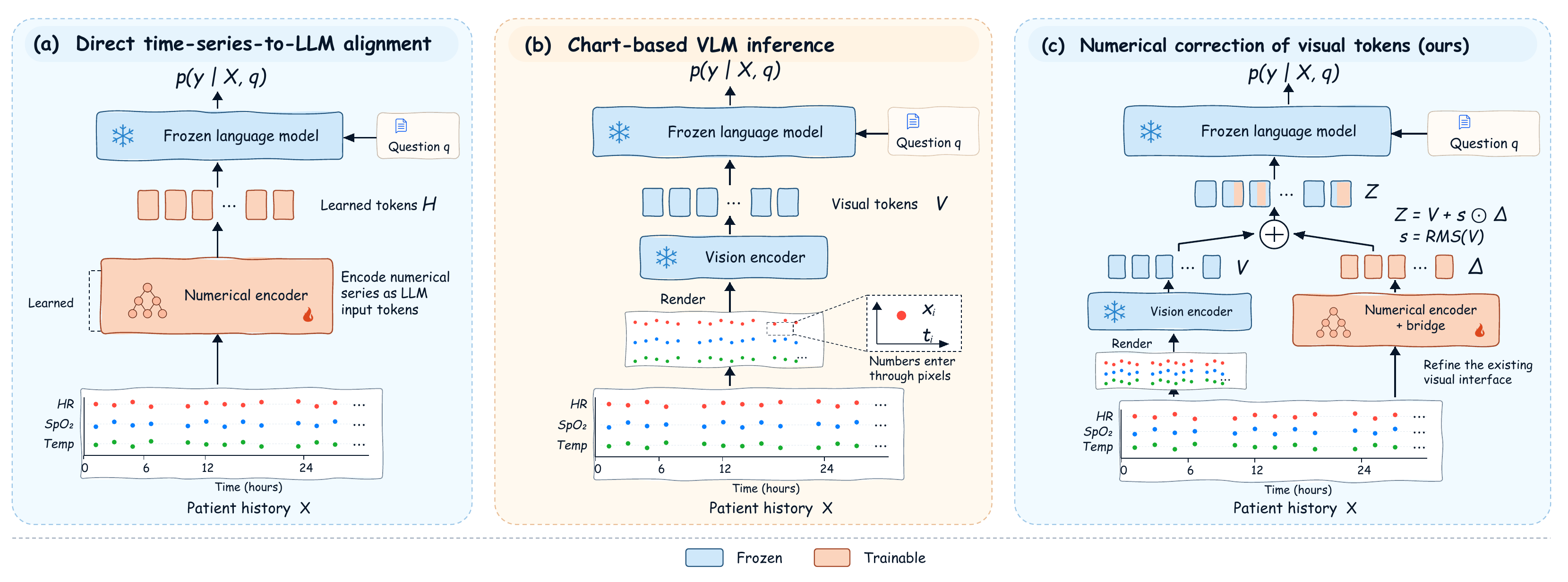}
\caption{\textbf{Numerical correction of an established visual interface.} Direct numerical interfaces learn how to present a new continuous representation to a language model. Chart inputs use its pretrained visual pathway. \mname{} adds numerical corrections to original visual tokens at their existing positions, preserving token order and positional indices. The language model and its output head remain frozen.}
\label{fig:concept}
\vspace{-1em}
\end{figure}

Our design starts from chart tokens that already pass through the pretrained visual-language interface. We add corrections derived from the underlying measurements at those same positions, learning from an established input pathway. Figure~\ref{fig:concept} summarizes the three input paradigms.

We introduce \mname{} (\textbf{Vi}sually aligned \textbf{S}eries-to-language \textbf{T}emporal \textbf{A}dapter), comprising an irregular-series encoder and a residual bridge. The encoder summarizes values, times, observation gaps, and variable identity; queries at visual positions read these summaries to correct visual tokens. Zero initialization recovers the original chart model. We train separate adapters for outcome prediction and temporal question answering (QA), and also evaluate QA after training only on outcomes.

Our contributions are threefold. First, we introduce a numerical adapter with fewer than one million trainable parameters that enriches the visual representations of a fully frozen vision-language model. Second, at 4 billion parameters, \mname{} reaches average precision of 0.4650 for acute kidney injury and 0.2651 for mortality, representing relative AP improvements of 3.15\% and 7.59\% over the strongest competing adaptation for each outcome, and 66.37\% and 77.56\% over the unadapted chart model, respectively. The 2-billion-parameter model's AKI AUROC is 0.7376, compared with GPT-5.6 Sol's 0.7380. Third, outcome-only training yields 41.73\% temporal QA accuracy at 4 billion parameters, 2.33 percentage points above native text (39.40\%). QA training yields 69.27\%, 29.87 percentage points above the same reference and 2.82-4.88 percentage points below the LoRA baselines, with more than 90\% fewer trainable parameters. We further observe consistent improvements in clinical prediction as the pretrained backbone scales from 2 to 9 billion parameters.

\section{Related Work}

\paragraph{Visual and multimodal time-series analysis.}
Charts expose temporal patterns through pretrained visual features, but recovering individual measurements depends on the rendering and visual representation \citep{images_signals}. CaTS-Bench finds that even strong vision-language models struggle with numerical nuances in time-series descriptions \citep{cats}. MLLM4TS aligns temporal patches with plot features for prediction \citep{mllm4ts}, but its task-specific output heads do not directly support a shared language interface for prediction and QA. \mname{} supplies numerical evidence through residuals at the original visual positions, allowing both tasks to use the unchanged language decoder and output head.

\paragraph{Connecting time series to language models.}
Time-LLM reprograms numerical patches using text prototypes and projects language-model representations into forecasts \citep{timellm}. ChatTS learns continuous representations using synthetic descriptions and questions; Time-MQA trains on contextualized numerical sequences for temporal QA \citep{chatts,timemqa}. ITFormer connects temporal encoders to frozen language models through instruction-aware interaction, while OpenTSLM explores soft prompts and gated cross-attention for medical signals \citep{itformer,opentslm}. Text requires sequences of values and timestamps; learned representations require alignment with the language model, even when its decoder is frozen. \mname{} starts from the existing chart representation, using a zero-initialized residual to learn how numerical evidence should modify this established input.

\paragraph{Parameter-efficient adaptation and clinical evaluation.}
Representation adaptation has supported knowledge transfer in biomedical prediction~\citep{shi2026deep}. LoRA adapts pretrained weights through low-rank updates \citep{lora}; BLIP-2 and Flamingo learn interfaces between pretrained components \citep{blip2,flamingo}. Parameter reduction alone does not determine access to numerical evidence. We study where to adapt: a small numerical module augments an existing visual interface while all pretrained components remain frozen. Clinical prediction remains challenging for general-purpose language models \citep{clinicrealm}, and temporal QA requires more than recognizing global trends \citep{clir}. We assess both tasks on the same backbone, including QA after outcome-only training, to examine task specialization and behavior beyond the training objective.

\section{\mname{}: Adding Numerical Information to Visual Tokens}
\label{sec:method}
\begin{figure}[t]
\centering
\includegraphics[width=\linewidth]{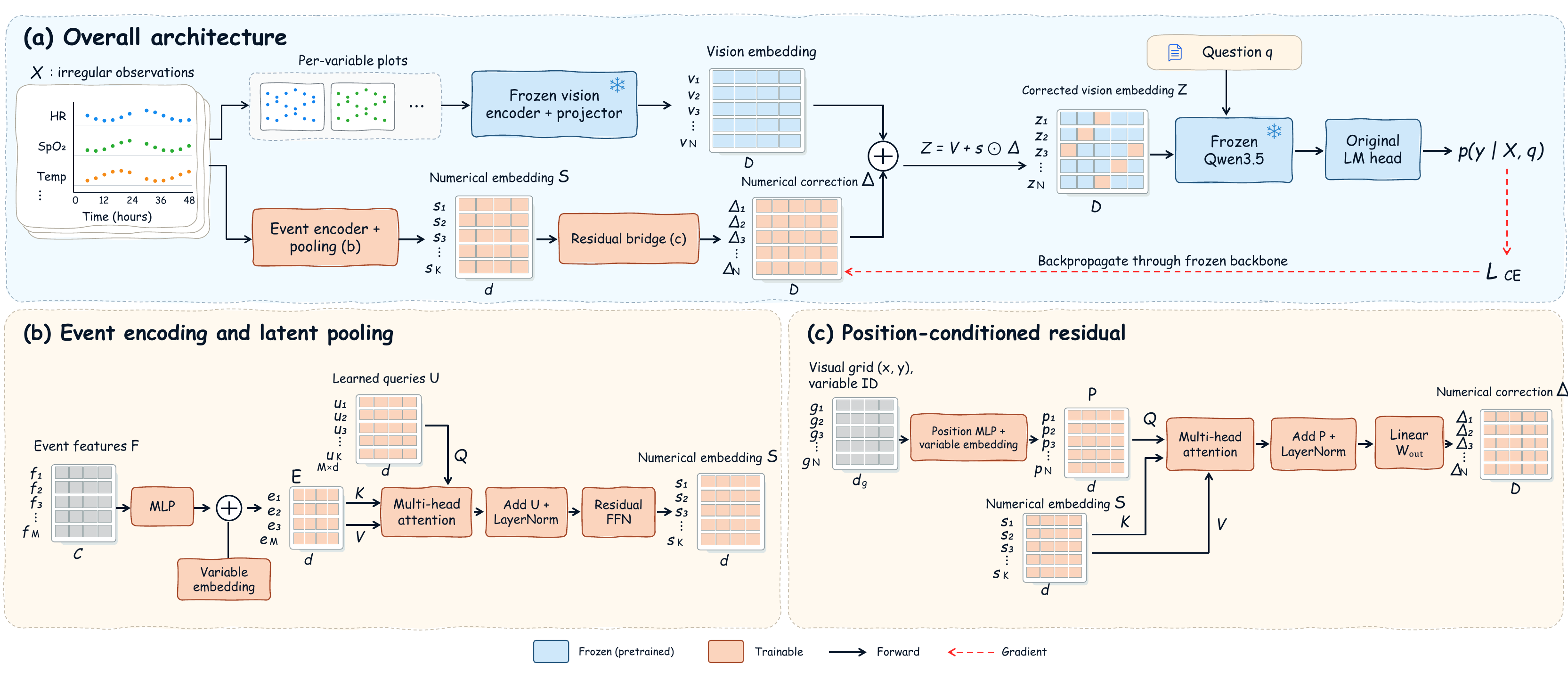}
\caption{\textbf{\mname{} architecture.} An event encoder pools irregular numerical observations into latent summaries. Queries derived from visual grid positions and variable identity read these summaries and produce residuals with the same shape as the original visual tokens. The residual output projection starts at zero. Answer supervision propagates through the frozen language model to the event encoder and bridge.}
\label{fig:architecture}
\end{figure}

Our design idea is simple and intuitive: charts provide an established visual input to a pretrained model, and their underlying measurements supply direct numerical evidence. \mname{} learns how this evidence should modify the existing visual tokens (Figure~\ref{fig:architecture}).
\subsection{Problem formulation and the pretrained visual pathway}
A patient history is a collection of irregular observations
$X=\{(t_i,c_i,x_i)\}_{i=1}^{M}$, where $t_i$ is the measurement time, $c_i\in\{1,\ldots,C\}$ identifies a clinical variable, and $x_i$ is its value. Each question specifies a visible interval and, for prospective tasks, an information-availability cutoff. We denote visible intervention context by $a$ and the question by $q$. The model represents $p(y\mid X,a,q)$ through its original language output head.

We render each variable as a separate chart image, with visible timestamps and values, and use the vision encoder and projector of the pretrained multi-modal LLM model, such as Qwen3.5, to obtain
\begin{equation}
 V=f_{\mathrm{vis}}(\operatorname{Render}(X))\in\R^{N\times D}.
\end{equation}
Here $N$ is the visual token count and $D$ the language model's hidden dimension. Retaining chart order, image delimiters, and positional indices lets numerical adaptation operate within the decoder's expected input structure.

\subsection{Encoding irregular numerical observations}
Encoding individual events exposes both measurements and observation gaps. For observation $i$, $\psi_i\in\R^{d_{\mathrm{in}}}$ contains its normalized value, relative time within the visible interval, time since the preceding measurement of that variable, and a first-observation indicator. A multilayer perceptron (MLP) and variable embedding produce
\begin{equation}
 e_i=f_\theta(\psi_i)+b_{c_i}\in\R^d.
\end{equation}
Unobserved variables receive a learned missing-observation token plus their variable embedding, explicitly representing absence. Stacking all tokens gives memory $E\in\R^{M_e\times d}$.

To summarize variable-length histories into a fixed-size memory, learned queries $U\in\R^{K\times d}$, indexed by variable, pool the events:
\begin{equation}
 R=\LN\!\left(U+\MHA(U,E,E)\right),\qquad
 S=R+\FFN(R).
 \label{eq:pool}
\end{equation}
MHA denotes multi-head attention, LN layer normalization, and FFN a feedforward network. Each summary can draw on events across time. \textbf{Joint} encoding lets queries attend to all variables to model their relationships during pooling. \textbf{Separate} encoding restricts query groups to their own variable, keeping evidence separate until the bridge. Both share attention and feedforward weights, have identical parameter counts, and expose all $K$ summaries to the bridge.

\subsection{Correcting original visual tokens}
To tailor each correction to its chart location, we condition retrieval on token $j$'s normalized grid coordinate $r_j$ and chart identity $c(j)$, using $p_j=h_\theta(r_j)+b_{c(j)}+W_vv_j$. Stacking queries into $P\in\R^{N\times d}$, the bridge retrieves numerical evidence in the visual dimension:
\begin{equation}
 \Delta=\operatorname{Linear}_{\mathrm{out}}\!\left[
 \LN\!\left(P+\MHA(P,S,S)\right)\right]\in\R^{N\times D}.
 \label{eq:bridge}
\end{equation}
Residual addition makes the visual feature the starting point for adaptation. We scale the correction relative to each token's magnitude:
\begin{equation}
 z_j=v_j+s_j\Delta_j,\qquad
 s_j=\sqrt{D^{-1}\textstyle\sum_{k=1}^{D}v_{jk}^{2}}.
 \label{eq:fusion}
\end{equation}
The scale $s_j$ is the root mean square (RMS) of the frozen visual token. For nonzero $v_j$, $\|s_j\Delta_j\|_2/\|v_j\|_2=\RMS(\Delta_j)$, so the adapter learns a relative correction magnitude. We initialize the output projection's weights and bias to zero, giving $Z=V$. The frozen decoder thus initially receives its native chart representation, and supervision subsequently determines the numerical correction.

Position-conditioned queries can retrieve across times and variables. The $K$ summaries remain internal, so the decoder still receives $N$ visual tokens. For fixed widths, interface attention scales with $M_eK+NK$.

\subsection{Learning through the original language interface}
Expressing predictions and QA responses as language answers gives both tasks the same output interface. With pretrained parameters $\phi$ frozen, we optimize the event encoder and bridge parameters $\theta$ using answer-token cross-entropy:
\begin{equation}
 \mathcal L(\theta)=-\sum_{\ell=1}^{|y|}
 \log p_\phi(y_\ell\mid y_{<\ell},q,a,Z_\theta(X)).
 \label{eq:objective}
\end{equation}
The vision encoder, pretrained projector, language decoder, embeddings, and language head remain frozen. Gradients through the decoder teach the adapter which numerical corrections are useful for the requested answer.

Binary outcomes use the original vocabulary tokens for the two labels, with $p(y=1)=\exp(\ell_1)/[\exp(\ell_0)+\exp(\ell_1)]$ from their logits. Multiple-choice QA normalizes over its answer labels in the same way. The question guides the decoder's use of the corrected record, allowing different questions about an unchanged visible interval to reuse its representation. The original output head also leaves autoregressive generation available; our experiments evaluate outcome prediction and temporal multiple-choice QA.

\section{Experimental Setup}
\label{sec:experiments}
\paragraph{Tasks and data.}
We use MIMIC-IV v3.1 \citep{mimiciv,mimiciv_data}, with eleven irregularly observed clinical variables and visible intervention records. Patient splits are disjoint. Observations from the first 24 hours in the intensive care unit (ICU) predict newly documented creatinine-defined acute kidney injury (AKI) during hours $(24,48]$ and in-hospital mortality. These outcomes have 5,000 training examples for each task, with 500 validation and 1,000 test examples each. Temporal QA contains 22,000 training, 1,100 validation, and 2,200 test questions across eleven categories following CLIR-Bench \citep{clir}, including event localization, cross-variable retrieval, trends, missingness, numerical summaries, and forecasting. Prospective questions use only information available at their cutoff. Appendices~\ref{app:data} and~\ref{app:qa} define the cohorts and tasks.

\paragraph{Prediction and question-answering training.}
We evaluate the multi-modal Qwen3.5 models with 2, 4, and 9 billion parameters, denoted 2B, 4B, and 9B \citep{qwen}. For prediction, one adapter is trained on both clinical outcomes and selected by validation average precision. We also evaluate this adapter on QA, without using QA examples or labels for training or model selection. For supervised QA, a separate adapter is trained on question-answer pairs and selected by validation accuracy. We evaluate both pooling designs defined in Section~\ref{sec:method}: Joint combines observations across variables, and Separate first summarizes each variable independently.

\paragraph{Comparisons.}
The unadapted models receive numerical text or time-series charts, labeled \emph{Unadapted text} and \emph{Unadapted plots}. Chart LoRA \citep{cats} and Text LoRA \citep{timemqa} follow the cited approaches to adapt the decoder using charts and serialized measurements, respectively. We also compare ChatTS, MLLM4TS, ITFormer, and OpenTSLM. Each implementation follows the original method, with input formats, interface dimensions, and task-specific output heads adjusted for the shared Qwen models and clinical tasks. All follow a common training and selection protocol. GPT-5.6 Sol-high~\citep{openai2026gpt56sol} and Luna-high~\citep{openai2026gpt56luna} provide zero-shot references; specialized RNN- and Transformer-based clinical predictors~\citep{ma2020concare,kim2024vita,wang2025colacare} appear in Appendix~\ref{app:references}.

Table~\ref{tab:adaptation_setup} summarizes decoder updates and trainable parameter counts. Among the trained adaptations, \mname{} uses the fewest trainable parameters: 90.61\% fewer than Chart LoRA and 91.73\% fewer than Text LoRA at 4B.
\begin{table}[!htbp]
\centering\small
\caption{\textbf{Comparison methods and adaptation size.} Update describes adaptation within the language decoder. Parameters are trainable counts at 4B. \mname{} uses 0.516M/0.582M/0.780M at 2B/4B/9B.}
\label{tab:adaptation_setup}
\setlength{\tabcolsep}{4pt}
\begin{tabular}{lcr}
\toprule
Method & Update & Trainable params (M)\\
\midrule
Unadapted text & None & 0\\
Unadapted plots & None & 0\\
\midrule
Chart LoRA & LoRA & 6.193\\
Text LoRA & LoRA & 7.032\\
ChatTS & LoRA & 6.692\\
MLLM4TS & Layer norm & 0.884\\
ITFormer & Frozen & 0.975\\
OpenTSLM & Cross-attention & 3.074\\
\midrule
\mname{}-Joint & Frozen & 0.582\\
\mname{}-Separate & Frozen & 0.582\\
\bottomrule
\end{tabular}
\end{table}

\paragraph{Training and model selection.}
For each backbone size, all compared methods use the same pretrained backbone, patient splits, training data, and input information. We use a consistent validation frequency, early-stopping rule, and best-checkpoint selection criterion across methods. Hyperparameters are selected by grid search for each method, and all configurations are determined exclusively on the validation set.

\paragraph{Metrics.}
Prediction uses area under the receiver operating characteristic curve (AUROC) and average precision (AP); QA uses answer accuracy. We assess predicted probabilities with Brier score and expected calibration error (ECE; ten equal-width bins; Appendix~\ref{app:calibration}). Tables report standard deviations (SD) from patient-level bootstrap resampling, keeping each patient's questions together. QA differences are expressed in percentage points; AUROC and AP differences are absolute changes on the 0-1 scale.

\section{Results and Analysis}
\label{sec:results}
We first compare outcome prediction and supervised temporal QA, then examine the effects of model size and question category.

\subsection{Compact adaptation improves outcome prediction}
\begin{table}[!t]
\centering\footnotesize
\caption{\textbf{Prediction and QA performance after outcome training.} Values are mean $\pm$ patient-bootstrap SD; bold and underline mark the best and second-best Qwen means per scale and metric. Adapted Qwen models use outcome supervision only. GPT models receive no task-specific training, and their QA scores appear in Table~\ref{tab:qa}.}
\label{tab:prediction}
\setlength{\tabcolsep}{1.3pt}
\begin{tabular}{lrrrrr}
\toprule
& \multicolumn{2}{c}{AKI} & \multicolumn{2}{c}{Mortality} & QA\\
\cmidrule(lr){2-3}\cmidrule(lr){4-5}
Method / input & AUROC & AP & AUROC & AP & Acc. (\%)\\
\midrule
\multicolumn{6}{l}{\emph{Proprietary references}}\\
GPT 5.6 Sol-high / text & 0.7380 $\pm 0.0293$ & 0.4212 $\pm 0.0246$ & 0.7515 $\pm 0.0231$ & 0.2935 $\pm 0.0254$ & -\\
GPT 5.6 Sol-high  / plots & 0.7283 $\pm 0.0292$ & 0.4126 $\pm 0.0330$ & 0.7221 $\pm 0.0291$ & 0.2733 $\pm 0.0285$ & -\\
GPT 5.6 Luna-high  / text & 0.7422 $\pm 0.0297$ & 0.4382 $\pm 0.0312$ & 0.6870 $\pm 0.0306$ & 0.2780 $\pm 0.0324$ & -\\
GPT 5.6 Luna-high  / plots & 0.6689 $\pm 0.0284$ & 0.3358 $\pm 0.0342$ & 0.6684 $\pm 0.0306$ & 0.2605 $\pm 0.0368$ & -\\
\midrule
\multicolumn{6}{l}{\emph{Qwen3.5-2B}}\\
Unadapted text & 0.4249 $\pm 0.0217$ & 0.1774 $\pm 0.0131$ & 0.5247 $\pm 0.0327$ & 0.1264 $\pm 0.0181$ & 32.85 $\pm 0.99$\\
Unadapted plots & 0.5297 $\pm 0.0223$ & 0.2301 $\pm 0.0196$ & 0.5235 $\pm 0.0297$ & 0.1172 $\pm 0.0154$ & 30.27 $\pm 0.98$\\
Chart LoRA & 0.7117 $\pm 0.0209$ & 0.3988 $\pm 0.0336$ & 0.6703 $\pm 0.0273$ & 0.1906 $\pm 0.0263$ & 31.28 $\pm 0.93$\\
Text LoRA & 0.7052 $\pm 0.0195$ & 0.3957 $\pm 0.0366$ & 0.6663 $\pm 0.0286$ & 0.1956 $\pm 0.0310$ & 32.16 $\pm 1.04$\\
ChatTS & 0.6297 $\pm 0.0220$ & 0.3193 $\pm 0.0299$ & 0.6673 $\pm 0.0257$ & 0.1890 $\pm 0.0277$ & 29.87 $\pm 0.93$\\
MLLM4TS & 0.6606 $\pm 0.0217$ & 0.3139 $\pm 0.0256$ & 0.6793 $\pm 0.0245$ & 0.1935 $\pm 0.0260$ & 27.67 $\pm 0.96$\\
ITFormer & 0.5958 $\pm 0.0220$ & 0.2869 $\pm 0.0287$ & 0.6141 $\pm 0.0301$ & 0.1767 $\pm 0.0244$ & 28.58 $\pm 0.89$\\
OpenTSLM & 0.5405 $\pm 0.0225$ & 0.2911 $\pm 0.0264$ & 0.5801 $\pm 0.0319$ & 0.1616 $\pm 0.0258$ & 29.50 $\pm 0.93$\\
\cmidrule(lr){1-6}
\mname{}-Joint & \textbf{0.7376} $\pm 0.0202$ & \textbf{0.4360} $\pm 0.0245$ & \textbf{0.6989} $\pm 0.0279$ & \textbf{0.2391} $\pm 0.0309$ & \textbf{33.27} $\pm 0.96$\\
\mname{}-Separate & \underline{0.7292} $\pm 0.0202$ & \underline{0.4261} $\pm 0.0246$ & \underline{0.6816} $\pm 0.0275$ & \underline{0.2047} $\pm 0.0295$ & \underline{33.18} $\pm 0.98$\\
\midrule
\multicolumn{6}{l}{\emph{Qwen3.5-4B}}\\
Unadapted text & 0.5124 $\pm 0.0213$ & 0.2078 $\pm 0.0173$ & 0.6352 $\pm 0.0301$ & 0.1962 $\pm 0.0321$ & 39.40 $\pm 1.08$\\
Unadapted plots & 0.6250 $\pm 0.0212$ & 0.2795 $\pm 0.0232$ & 0.5872 $\pm 0.0288$ & 0.1493 $\pm 0.0211$ & 35.50 $\pm 0.95$\\
Chart LoRA & 0.7292 $\pm 0.0203$ & \underline{0.4508} $\pm 0.0354$ & 0.7156 $\pm 0.0259$ & 0.2285 $\pm 0.0335$ & 40.95 $\pm 0.95$\\
Text LoRA & 0.7279 $\pm 0.0203$ & 0.4412 $\pm 0.0386$ & \underline{0.7256} $\pm 0.0270$ & \underline{0.2464} $\pm 0.0372$ & \textbf{42.05} $\pm 1.12$\\
ChatTS & 0.6801 $\pm 0.0217$ & 0.3624 $\pm 0.0314$ & 0.7076 $\pm 0.0243$ & 0.2347 $\pm 0.0350$ & 34.68 $\pm 0.97$\\
MLLM4TS & 0.6952 $\pm 0.0208$ & 0.3468 $\pm 0.0283$ & 0.7161 $\pm 0.0235$ & 0.2372 $\pm 0.0339$ & 30.91 $\pm 0.98$\\
ITFormer & 0.6591 $\pm 0.0219$ & 0.3347 $\pm 0.0298$ & 0.6660 $\pm 0.0268$ & 0.2029 $\pm 0.0316$ & 31.05 $\pm 0.96$\\
OpenTSLM & 0.6324 $\pm 0.0232$ & 0.3291 $\pm 0.0285$ & 0.6367 $\pm 0.0291$ & 0.1908 $\pm 0.0307$ & 30.82 $\pm 0.98$\\
\cmidrule(lr){1-6}
\mname{}-Joint & \textbf{0.7440} $\pm 0.0202$ & \textbf{0.4650} $\pm 0.0262$ & \textbf{0.7350} $\pm 0.0258$ & \textbf{0.2651} $\pm 0.0384$ & \underline{41.73} $\pm 1.00$\\
\mname{}-Separate & \underline{0.7378} $\pm 0.0208$ & 0.4432 $\pm 0.0262$ & 0.7238 $\pm 0.0264$ & 0.2392 $\pm 0.0226$ & 41.64 $\pm 1.01$\\
\midrule
\multicolumn{6}{l}{\emph{Qwen3.5-9B}}\\
Unadapted text & 0.4046 $\pm 0.0206$ & 0.1313 $\pm 0.0170$ & 0.6104 $\pm 0.0295$ & 0.1950 $\pm 0.0374$ & 47.00 $\pm 1.05$\\
Unadapted plots & 0.5204 $\pm 0.0220$ & 0.2115 $\pm 0.0186$ & 0.5643 $\pm 0.0290$ & 0.1430 $\pm 0.0204$ & 45.09 $\pm 0.99$\\
Chart LoRA & 0.7377 $\pm 0.0206$ & \underline{0.4652} $\pm 0.0349$ & 0.7451 $\pm 0.0256$ & 0.2582 $\pm 0.0378$ & 43.22 $\pm 0.97$\\
Text LoRA & 0.7343 $\pm 0.0203$ & 0.4521 $\pm 0.0363$ & 0.7430 $\pm 0.0258$ & 0.2777 $\pm 0.0421$ & 43.67 $\pm 1.07$\\
ChatTS & 0.6476 $\pm 0.0225$ & 0.3329 $\pm 0.0302$ & 0.7416 $\pm 0.0231$ & 0.2726 $\pm 0.0380$ & 44.28 $\pm 0.97$\\
MLLM4TS & 0.6709 $\pm 0.0211$ & 0.3030 $\pm 0.0299$ & \underline{0.7468} $\pm 0.0231$ & \underline{0.2796} $\pm 0.0386$ & 42.58 $\pm 0.90$\\
ITFormer & 0.5935 $\pm 0.0223$ & 0.3001 $\pm 0.0304$ & 0.6811 $\pm 0.0267$ & 0.2313 $\pm 0.0364$ & 42.62 $\pm 0.92$\\
OpenTSLM & 0.5374 $\pm 0.0245$ & 0.2679 $\pm 0.0296$ & 0.6300 $\pm 0.0283$ & 0.2060 $\pm 0.0346$ & 42.46 $\pm 0.95$\\
\cmidrule(lr){1-6}
\mname{}-Joint & \textbf{0.7502} $\pm 0.0203$ & \textbf{0.4840} $\pm 0.0266$ & \textbf{0.7559} $\pm 0.0256$ & \textbf{0.3021} $\pm 0.0327$ & \underline{51.00} $\pm 0.97$\\
\mname{}-Separate & \underline{0.7423} $\pm 0.0201$ & 0.4627 $\pm 0.0259$ & 0.7440 $\pm 0.0249$ & 0.2646 $\pm 0.0276$ & \textbf{51.73} $\pm 0.96$\\
\bottomrule
\end{tabular}
\end{table}

\paragraph{Prediction through a frozen backbone.}
\mname{}-Joint has the highest mean scores on all four prediction metrics at every Qwen scale (Table~\ref{tab:prediction}). At 4B, AKI AP is 0.4650 versus Chart LoRA's 0.4508, an absolute difference of 0.0142; mortality AP is 0.2651 versus Text LoRA's 0.2464, a difference of 0.0187. Against native charts, the corresponding absolute differences are 0.1855 and 0.1158. These gains use the smallest trainable interface.

\paragraph{Probability quality also improves.}
\label{sec:calibration}
Native models must predict future outcomes from irregular histories without endpoint-specific training. Their weak ranking, including sub-chance AKI AUROC at 2B and 9B, illustrates the difficulty of applying untuned language models to structured clinical prediction \citep{clinicrealm}. Their normalized label probabilities can also reflect answer preferences \citep{contextual_calibration}, so accurate risk calibration is not inherent to the language interface. After adaptation, Brier score improves over native charts in all twelve combinations of size, outcome, and encoding, and ECE in ten, without post-hoc calibration (Appendix~\ref{app:calibration}). At 9B, mortality ECE falls from 0.7261 to 0.0666 with \mname{}-Joint and 0.0284 with \mname{}-Separate.

\paragraph{Prediction gains do not automatically improve downstream QA.}
At 9B, all six alternatives have lower mean QA accuracy than native charts. Chart LoRA is lower by 1.87 percentage points against native charts and Text LoRA by 3.33 against native text. Outcome supervision may favor features or answer-token preferences for binary prediction over temporal questions. Frozen-decoder ITFormer also scores below both references, suggesting that input adaptation itself can affect QA. \mname{}-Joint's mean accuracy is higher than native text by 0.42, 2.33, and 4.00 percentage points across scales. At 4B, Text LoRA scores 42.05\% versus \mname{}-Joint's 41.73\%, a difference of 0.32 percentage points.

\paragraph{Comparison with GPT-5.6 Sol.}
The 2B \mname{}-Joint model's AKI AUROC is 0.7376 versus 0.7380 for text-input Sol; its AP is 0.4360 versus 0.4212. At 9B, \mname{}-Joint has higher mean scores across prediction metrics and both Sol input formats. Clinical adaptation brings compact models to this prediction level.

\FloatBarrier
\subsection{Strong supervised QA with a compact numerical interface}
\begin{table}[!t]
\centering\footnotesize
\caption{\textbf{Large QA gains with a compact frozen interface.} Accuracy (\%) is mean $\pm$ patient-bootstrap SD; $\Delta$ is the gain over the unadapted chart model in percentage points. GPT models receive no task-specific training. Qwen adaptations are trained on QA; bold and underline mark the best and second-best Qwen means per scale.}
\label{tab:qa}
\setlength{\tabcolsep}{2pt}
\begin{tabular}{@{}lrrrrrr@{}}
\toprule
GPT reference / input & \multicolumn{6}{c}{QA without task-specific training (\%)}\\
\midrule
GPT 5.6 Sol-high / text & \multicolumn{6}{c}{90.51 $\pm 0.70$}\\
GPT 5.6 Sol-high / plots & \multicolumn{6}{c}{84.24 $\pm 0.79$}\\
GPT 5.6 Luna-high / text & \multicolumn{6}{c}{80.30 $\pm 0.93$}\\
GPT 5.6 Luna-high / plots & \multicolumn{6}{c}{74.05 $\pm 1.13$}\\
\midrule
Method & \multicolumn{2}{c}{2B} & \multicolumn{2}{c}{4B} & \multicolumn{2}{c}{9B}\\
\cmidrule(lr){2-3}\cmidrule(lr){4-5}\cmidrule(lr){6-7}
& Acc. (\%) & $\Delta$ (pp) & Acc. (\%) & $\Delta$ (pp) & Acc. (\%) & $\Delta$ (pp)\\
\midrule
Unadapted text & 32.85 $\pm 0.99$ & +2.58 & 39.40 $\pm 1.08$ & +3.90 & 47.00 $\pm 1.05$ & +1.91\\
Unadapted plots & 30.27 $\pm 0.98$ & +0.00 & 35.50 $\pm 0.95$ & +0.00 & 45.09 $\pm 0.99$ & +0.00\\
\midrule
Chart LoRA & \underline{67.42} $\pm 0.87$ & +37.15 & \underline{72.09} $\pm 0.92$ & +36.59 & \underline{73.00} $\pm 0.88$ & +27.91\\
Text LoRA & \textbf{67.78} $\pm 0.94$ & +37.51 & \textbf{74.15} $\pm 0.94$ & +38.65 & \textbf{75.04} $\pm 0.93$ & +29.95\\
ChatTS & 54.12 $\pm 1.07$ & +23.85 & 59.18 $\pm 1.02$ & +23.68 & 62.81 $\pm 1.01$ & +17.72\\
MLLM4TS & 52.66 $\pm 1.00$ & +22.39 & 58.50 $\pm 1.02$ & +23.00 & 62.05 $\pm 0.99$ & +16.96\\
ITFormer & 52.31 $\pm 1.01$ & +22.04 & 57.36 $\pm 1.05$ & +21.86 & 62.00 $\pm 0.97$ & +16.91\\
OpenTSLM & 51.72 $\pm 0.99$ & +21.45 & 57.41 $\pm 1.03$ & +21.91 & 62.05 $\pm 0.99$ & +16.96\\
\midrule
\mname{}-Joint & 63.05 $\pm 1.00$ & +32.78 & 68.27 $\pm 0.99$ & +32.77 & 70.55 $\pm 0.95$ & +25.46\\
\mname{}-Separate & 63.55 $\pm 0.97$ & +33.28 & 69.27 $\pm 0.96$ & +33.77 & 69.32 $\pm 0.93$ & +24.23\\
\bottomrule
\end{tabular}
\end{table}

Chart LoRA and Text LoRA achieve the highest mean supervised QA accuracy among Qwen adaptations at every scale (Table~\ref{tab:qa}). At 4B, they reach 72.09\% and 74.15\%, versus 69.27\% for \mname{}-Separate. LoRA directly adapts the decoder to questions and answers, giving it greater task-specific flexibility. \mname{} learns how to present numerical evidence to a fixed decoder.

At 4B, \mname{}-Separate gains 29.87 percentage points over native text and 33.77 over native charts, coming within 2.82-4.88 points of LoRA with more than 90\% fewer trainable parameters. Both encodings outperform ChatTS, MLLM4TS, ITFormer, and OpenTSLM across scales. The contribution is strong QA through a compact interface, using the same frozen-backbone design that leads the prediction comparisons.

GPT-5.6 Sol remains substantially stronger: 90.51\% with text and 84.24\% with charts, without task-specific training. This highlights the temporal reasoning capabilities of stronger proprietary models. Numerical adaptation narrows the gap for compact open models, motivating its evaluation with stronger pretrained backbones.

\FloatBarrier
\subsection{Larger backbones further improve numerical adaptation}
\label{sec:scaling}
\begin{figure}[!htbp]
\centering
\includegraphics[width=\linewidth]{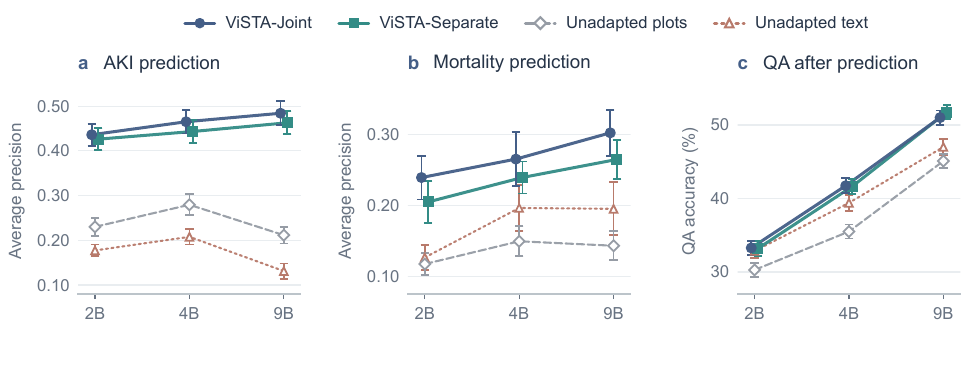}
\caption{\textbf{Prediction gains persist as the backbone grows.} Adapted models use outcome supervision only; the right panel evaluates QA without QA training. Error bars show patient-bootstrap SD. Both encodings have identical parameter counts at each scale.}
\label{fig:scaling}
\end{figure}

\paragraph{Prediction improves through 9B.}
\mname{}-Joint AKI AP rises from 0.4360 to 0.4840 and mortality AP from 0.2391 to 0.3021 across 2B-9B (Figure~\ref{fig:scaling}). Native prediction lacks this consistent trend; adapted 2B even exceeds native 9B charts on both AP measures (0.2115 and 0.1430). Capacity benefits prediction when coupled with numerical adaptation.

\paragraph{Early cross-variable pooling benefits prediction more than QA.}
\mname{}-Joint has higher mean scores than \mname{}-Separate on every prediction metric, while outcome-trained QA differs by at most 0.73 percentage points. Both bridges access all variables; joint pooling additionally mixes them before compression. This suggests early interaction helps retain combinations relevant to patient risk. Supervised QA favors \mname{}-Separate at 2B and 4B, \mname{}-Joint at 9B. \mname{}-Separate's stronger 4B missingness counting (63.0\% versus 54.0\%) is consistent with preserving variable-specific detail. Pooling preference thus depends on the task (Appendix~\ref{app:encoding}).

\paragraph{Design ablations.}
Removing zero initialization or RMS scaling reduces all four prediction metrics and QA performance (Table~\ref{tab:ablation}). Keeping only visual tokens or only numerical residuals at inference lowers both prediction and QA scores. Appendix~\ref{app:ablation} discusses their interpretation.

\subsection{Supervised QA gains span numerical and temporal operations}
\begin{figure}[!htbp]
\centering
\includegraphics[width=\linewidth]{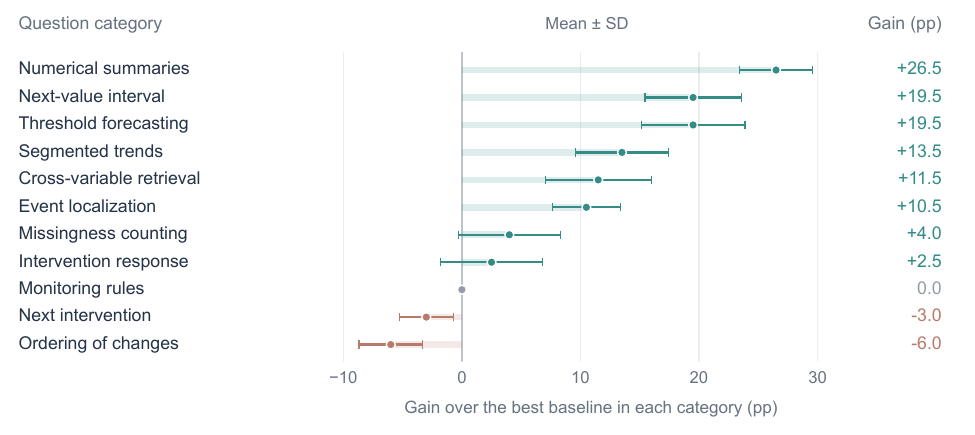}
\caption{\textbf{Supervised QA gains across most categories.} At 4B, \mname{}-Separate is compared with the category-wise best of ChatTS, MLLM4TS, ITFormer, and OpenTSLM. Error bars show paired patient-bootstrap SD with baseline selection repeated per replicate. Absolute scores appear in Appendix~\ref{app:categories}.}
\label{fig:categories}
\end{figure}

At 4B, \mname{}-Separate scores higher than the category-wise best of ChatTS, MLLM4TS, ITFormer, and OpenTSLM in eight categories, ties in one, and trails in two (Figure~\ref{fig:categories}). Gains span numerical summaries (+26.5 percentage points), threshold forecasting (+19.5), segmented trends (+13.5), and cross-variable retrieval (+11.5). These gains span numerical access and temporal operations under QA supervision.

\paragraph{The advantage persists without numerical summaries.}
Without numerical summaries, \mname{}-Separate averages 66.45\% across ten categories, versus 59.25\% for the category-wise best baseline. The 7.20 percentage-point advantage also holds over every individual baseline. Ordering of changes and next-intervention prediction remain weaker by 6.0 and 3.0 percentage points, respectively.

\section{Discussion and Conclusion}
\mname{} improves clinical prediction and supervised temporal QA with fewer than one million trainable parameters while keeping the pretrained model fixed. The 2B model approaches GPT-5.6 Sol on AKI prediction, prediction improves through 9B, and QA gains extend beyond numerical summaries. These results motivate applying the adapter to more capable pretrained models.

The evaluation uses one hospital system, one training seed, and automatically generated multiple-choice questions. Specialized clinical predictors and LoRA trained on QA remain stronger in their respective tasks. Matched training controls are needed to attribute QA gains to outcome supervision and isolate the contribution of visual features. Further evaluation should include external cohorts, open-ended interaction, and systems combining separate prediction and QA models~\citep{shi2025out}.

\clearpage
\subsection*{AI use statement}
Generative AI tools assist with language editing, manuscript organization, and numerical consistency checks. Methods and experiments are designed and executed by humans.

\subsection*{Ethics statement}
This study uses deidentified clinical records from MIMIC-IV under its controlled-access framework. The performance of OpenAI models on all datasets was processed using the secure Azure OpenAI API, with human review of the data waived. The experiments evaluate retrospective prediction and record understanding.

\subsection*{Reproducibility statement}
Section~\ref{sec:method} specifies the model, and Section~\ref{sec:experiments} describes the comparison methods. Appendices~\ref{app:data}-\ref{app:calibration} document cohort construction, question definitions, prompt templates, input visibility, implementation settings, and metric computation. Methods share a common validation and checkpoint-selection protocol, with hyperparameters chosen by grid search on the validation set. Reproducing the clinical data extraction requires authorized access to MIMIC-IV under its data-use agreement.

\bibliography{references}
\bibliographystyle{preprint}
\clearpage
\appendix
\makeatletter
\setlength{\@fptop}{0pt}
\setlength{\@fpsep}{18pt}
\setlength{\@fpbot}{0pt plus 1fil}
\makeatother
\section{Cohort and Outcome Definitions}
\label{app:data}
\paragraph{Study population.}
We include adults with an ICU stay of at least 24 hours in MIMIC-IV v3.1 and use the first eligible stay per patient. Training, validation, and test sets are disjoint at the patient level. A patient may contribute to both prediction endpoints or to multiple QA categories within the same split.

\paragraph{Clinical variables and availability.}
The eleven numerical variables are heart rate, mean arterial pressure, systolic pressure, diastolic pressure, oxygen saturation, respiratory rate, temperature, lactate, creatinine, white blood cell count, and urine output. Observations retain their values, variable identities, and timestamps. Prediction and prospective QA use only observations measured and available by the question cutoff. Visible intervention starts are provided as timestamped text; future observations and outcome labels remain hidden.

\paragraph{Acute kidney injury.}
The landmark is ICU hour 24 and the outcome window is $(24,48]$ hours. AKI is defined by a creatinine increase of at least 0.3 mg/dL within 48 hours or at least 1.5 times the minimum qualifying prior measurement within seven days. Eligibility requires an ICU stay of at least 48 hours, two baseline creatinine measurements available by the landmark (including one at or after hour 12), and follow-up measurement at or after hour 36. Patients with AKI or renal replacement therapy before the landmark are ineligible. This is a recorded creatinine-defined endpoint; urine-output staging is not included. Test prevalence is 20.5\%.

\paragraph{In-hospital mortality.}
The mortality cohort consists of patients alive and hospitalized at ICU hour 24. The outcome is death before discharge from the current hospitalization. Test prevalence is 10.9\%.

\section{Temporal Question Construction}
\label{app:qa}
The question taxonomy follows CLIR-Bench \citep{clir}. We could not reproduce the original benchmark inputs from the released materials. We therefore rebuilt questions, labels, and distractors on our MIMIC-IV cohort using an independent implementation of the published task definitions and answer-construction rules. The resulting evaluation uses our patient splits and newly generated questions. Each question presents four distinct answer options in randomized order. Table~\ref{tab:qa_definitions} defines the eleven categories. Appendix~\ref{app:prompts} gives prompt templates, and Appendix~\ref{app:qa_simple_baselines} reports category-wise majority-answer and random baselines.

\begin{table}[!htbp]\centering\small
\caption{\textbf{Temporal QA definitions.} Each question specifies its observation window, thresholds, and tolerances.}
\label{tab:qa_definitions}
\begin{tabularx}{\linewidth}{>{\raggedright\arraybackslash}p{0.25\linewidth}>{\raggedright\arraybackslash}X}\toprule
Category & Answer construction\\\midrule
TG: Event localization & Locate the start of the first consecutive pair of observed threshold breaches within one of four quarters of a time window.\\
ASR: Cross-variable retrieval & Find an extremum of one variable and retrieve the nearest observation of another; resolve time ties using the earlier reading.\\
TPR: Segmented trends & Split ordered observations into five nearly equal blocks, compare medians using a stated tolerance, and merge adjacent repeated directions.\\
MA: Missingness counting & Count a variable's observed and absent entries on the union of recorded timestamps in a specified window.\\
TSS: Numerical summaries & Select the correct first, last, minimum, and maximum values for two variables from summaries with numerical distractors.\\
CVR: Ordering of changes & Compare the first specified directional changes of two variables, including simultaneous and absent-change cases.\\
IR: Intervention response & Compare pre- and post-intervention medians within stated windows, requiring sufficient observations on both sides.\\
TF: Threshold forecasting & Predict a future observed threshold breach and identify the direction of the final two visible observations.\\
NIF: Next-value interval & Predict the interval containing the first subsequent observed value within a stated horizon.\\
IID: Next intervention & Predict the category combination at the first subsequent recorded intervention start.\\
MED: Monitoring rules & Apply a stated rule combining repeated blood-pressure and oxygen-saturation threshold breaches and observation sufficiency.\\\bottomrule
\end{tabularx}\end{table}

\paragraph{Retrospective and prospective questions.}
Retrospective questions concern the supplied observation window. Forecasting and next-intervention questions expose only information available at their cutoff. Intervention prediction targets recorded care, while intervention-response questions describe observed changes around treatment starts.

\paragraph{Answer construction.}
Thresholds and stability tolerances are stated in each question. Missingness counts absent entries on the union of observation timestamps. Numerical-summary distractors perturb values from the correct summary. These operations produce reproducible labels for evaluating temporal and numerical understanding.

\clearpage
\section{Prompt Templates}
\label{app:prompts}
The following templates show the common Qwen instructions and six task questions. Braced fields are filled from the visible record or question parameters. Chart inputs and numerical text use the same task wording and answer labels. Charts and intervention context contain only information available within the question's visible interval.

\begingroup
\newcommand{\promptbox}[2]{\par\medskip\noindent\textbf{#1}\par\nopagebreak\smallskip\noindent\fbox{\begin{minipage}{\dimexpr\linewidth-2\fboxsep-2\fboxrule\relax}\small\ttfamily\raggedright #2\end{minipage}}\par\medskip}

\subsection{Shared instructions and input format}
\promptbox{System message}{Answer the clinical record question using the supplied observations.}

\promptbox{Chart input: user message}{\{variable name 1\}\par
\{chart image 1\}\par
\ldots\par
\{variable name 11\}\par
\{chart image 11\}\par
Intervention starts: \{time\}: \{intervention\}; \ldots\par
\{task question and answer instruction\}}

\promptbox{Numerical text: user message}{Visible ICU history: \{start\} to \{end\} hours.\par
\{variable name\} (\{unit\}), hour:value: \{time\}:\{value\}; \ldots\par
\ldots\par
Intervention starts: \{time\}: \{intervention\}; \ldots\par
\{task question and answer instruction\}}
The numerical-text block repeats for all eleven variables; a variable with no measurements is represented by \texttt{No observations.} The chart format shows the per-variable rendering used in the experiments.

\subsection{Clinical prediction}
\promptbox{Acute kidney injury}{Using only information available by ICU hour 24: Will a new recorded creatinine-defined AKI event occur during ICU hours (24, 48]? An event is a creatinine rise of at least 0.3 mg/dL over a prior value within 48 hours or at least 1.5 times a prior value within 7 days. Answer 0 for no or 1 for yes.\par
Return only 0 or 1.}

\promptbox{In-hospital mortality}{Using only information available by ICU hour 24, will this patient die before discharge from the current hospitalization? Answer 0 for survival to discharge or 1 for in-hospital death.\par
Return only 0 or 1.}

\subsection{Temporal question answering}
Each QA question below is followed by the same option block. The four distinct answer texts are permuted before assigning letters; the correct letter depends on this permutation.
\promptbox{QA option block and answer instruction}{Options:\par
A. \{option A\}\par
B. \{option B\}\par
C. \{option C\}\par
D. \{option D\}\par
Return only the answer letter A, B, C, or D.}

\promptbox{Event localization}{Within ICU hours 0-\{end\}, which quarter of the window contains the start of the first pair of consecutive observed \{variable\} values both \{below or above\} \{threshold\} \{unit\}? Count observed pairs, without assuming values between measurements.}
The options identify the four quarters of the observation window, including their time boundaries.

\promptbox{Cross-variable retrieval}{Within ICU hours 0-\{end\}, find the earliest \{maximum or minimum\} of \{variable 1\}. Which interval contains the closest measured \{variable 2\} (\{unit\})? Resolve equal time distances using the earlier reading. Intervals include the lower endpoint only.}
The options are four numerical intervals for the second variable.

\promptbox{Numerical summaries}{Which numerical handoff summary matches the observed measurements during ICU hours 0-\{end\}?}
Each option lists the first, last, minimum, and maximum measurements for two variables, with their units. Distractors perturb these summary values.

\promptbox{Threshold forecasting}{Using only observations through ICU hour \{cutoff\}, predict whether any observed \{variable\} will be \{below or above\} \{threshold\} \{unit\} during the next \{horizon\} hours, and identify the direction of the last two visible readings.}
The options combine future threshold breach or absence with nondecreasing or decreasing last observed values. Only measurements available by the cutoff enter the input; subsequent observations determine the reference answer.
\endgroup

\clearpage

\section{Implementation and Evaluation Details}
\label{app:implementation}
\paragraph{Numerical features.}
Variable values are centered by the training-set median and scaled by the training interquartile range, floored at $10^{-3}$. For normalized value $\tilde x_i$, interval $[t_s,t_e]$, and within-variable gap $g_i$, the event features are
\begin{equation}
 \begin{aligned}
 \psi_i=\biggl[
 \operatorname{sign}(\tilde x_i)\log(1+|\tilde x_i|),\;
 \frac{t_i-t_s}{\max(t_e-t_s,1)},\;
 \frac{\log(1+\max(t_i,0))}{10},\\
 \qquad
 \frac{\log(1+g_i)}{10},\;
 \mathbf{1}_{\mathrm{first},i},\;
 \frac{\log(1+\max(t_e-t_s,1))}{10}
 \biggr].
 \end{aligned}
\end{equation}
Gaps and first-observation flags are recomputed using only visible observations. The event MLP has two linear layers with a SiLU activation. Missing-variable tokens are appended to the event memory, which accounts for the distinction between $M$ and $M_e$.

\begin{table}[!htbp]\centering\small
\caption{\textbf{Selected \mname{} interface and training configuration.} Settings are selected on the validation set.}
\label{tab:config}
\begin{tabular}{ll}\toprule
Setting & Value\\\midrule
Clinical variables $C$; slots per variable & 11; 12\\
Total internal slots $K$ & 132\\
Visual tokens per chart; total $N$ & 192; 2,112\\
Numerical optimizer & AdamW, learning rate $10^{-4}$\\
Weight decay; gradient norm ceiling & 0.01; 1.0\\
Numerical precision & BF16 backbone\\
Hyperparameter selection & Grid search on the validation set\\\bottomrule
\end{tabular}\end{table}

\paragraph{Positional queries and frozen components.}
Position features are normalized horizontal and vertical coordinates of the visual grid. A two-layer MLP maps coordinates to numerical width, and the chart's variable embedding is added. Residuals preserve image delimiters and visual positions. The final linear layer, including its bias, starts at zero. Pretrained parameters remain frozen, while gradients propagate through the language decoder to the interface.

\paragraph{Hyperparameter and model selection.}
All methods follow the protocol in Section~\ref{sec:experiments}. Hyperparameters are selected by grid search on the validation set. Prediction models are selected by validation AP and QA models by validation accuracy. The selected prediction model is also evaluated on QA without further training or QA-based selection. Test results do not influence configuration choices.

\paragraph{Input representations.}
Numerical text lists variable names, units, and timestamp-value observations. Visual inputs render one chart per variable. Both representations describe the history available at the question cutoff, together with visible intervention context.

\paragraph{Scoring and uncertainty.}
QA selects the most probable answer label. Prediction metrics use continuous positive-label probabilities obtained by normalizing the original language-head logits over the two binary labels. Training uses answer-token cross-entropy.

Standard deviations use 2,000 patient-level bootstrap resamples, keeping all questions from the same patient together. QA accuracy is averaged across categories. For paired differences, both methods use the same resampled patients. Comparisons with the best baseline in each category reselect that baseline within every resample. Bootstrap SD measures test-patient uncertainty for one trained seed.



\section{Scaling and Variable Encoding}
\label{app:scaling}
Figure~\ref{fig:scaling} traces prediction and QA performance after outcome training across scales, alongside both unadapted input baselines. Table~\ref{tab:scale} compares \mname{}-Joint's QA accuracy with the unadapted text baseline and the unadapted chart baseline. The respective differences are 0.42/2.33/4.00 and 3.00/6.23/5.91 percentage points at 2B, 4B, and 9B. The table separately reports supervised QA gains over the unadapted chart baseline. Interface size is 0.516M at 2B, 0.582M at 4B, and 0.780M at 9B.
\begin{table}[!htbp]
\centering\footnotesize
\caption{\textbf{Prediction and QA gains over the unadapted models.} Prediction AUROC changes use the unadapted chart baseline as the reference. QA after prediction training is compared with both unadapted input baselines. These columns use \mname{}-Joint; supervised QA gains use their respective encodings and are relative to the unadapted chart baseline.}
\label{tab:scale}
\setlength{\tabcolsep}{4pt}
\begin{tabular}{lrrrrrr}
\toprule
& \multicolumn{2}{c}{Prediction $\Delta$AUROC} & \multicolumn{2}{c}{QA after prediction (pp)} & \multicolumn{2}{c}{Supervised QA $\Delta$ (pp)}\\
\cmidrule(lr){2-3}\cmidrule(lr){4-5}\cmidrule(lr){6-7}
Scale & AKI & Mortality & vs. plots & vs. text & \mname{}-Joint & \mname{}-Separate\\
\midrule
2B & +0.2079 & +0.1754 & +3.00 & +0.42 & +32.78 & +33.28\\
4B & +0.1190 & +0.1478 & +6.23 & +2.33 & +32.77 & +33.77\\
9B & +0.2298 & +0.1916 & +5.91 & +4.00 & +25.46 & +24.23\\
\bottomrule
\end{tabular}
\end{table}

\subsection{Encoding comparison}
\label{app:encoding}
\mname{}-Joint and \mname{}-Separate differ in the attention mask during the first numerical pooling stage. Parameter sharing, latent width, slot count, and the downstream bridge are unchanged. \mname{}-Separate pools observations within each variable; subsequent visual queries can read all pooled variables. Table~\ref{tab:encoding} calculates \mname{}-Joint minus \mname{}-Separate from the final prediction and QA scores. \mname{}-Joint has higher mean prediction AP at every size. QA after prediction training differs by 0.09 percentage points at 2B and 4B, and favors \mname{}-Separate by 0.73 percentage points at 9B.
\begin{table}[!htbp]
\centering\small
\caption{\textbf{\mname{}-Joint minus \mname{}-Separate encoding.} Positive entries favor \mname{}-Joint. Prediction and supervised QA use separately trained adapters. The first QA column evaluates the prediction-trained adapter without further training.}
\label{tab:encoding}
\setlength{\tabcolsep}{4pt}
\begin{tabular}{lrrrr}
\toprule
Scale & AKI AP & Mortality AP & QA after prediction (pp) & Supervised QA (pp)\\
\midrule
2B & +0.0099 & +0.0344 & +0.09 & -0.50\\
4B & +0.0218 & +0.0259 & +0.09 & -1.00\\
9B & +0.0213 & +0.0375 & -0.73 & +1.23\\
\bottomrule
\end{tabular}
\end{table}

\section{Input Interventions and Design Ablations}
\label{app:ablation}
Table~\ref{tab:ablation} separates changes to the input of a trained model from changes to adapter training. All results use the 4B backbone and outcome supervision; QA uses no question-answer training examples.
\begin{table}[!htbp]
\centering\small
\caption{\textbf{Input interventions and design ablations at 4B.} The full model is \mname{}-Joint from Table~\ref{tab:prediction}. Here $R=s\odot\Delta$ denotes the scaled numerical residual. Visual-only and residual-only inputs are evaluated at inference without retraining. The last two rows remove the indicated component during training. QA is evaluated without QA supervision; higher is better for every metric.}
\label{tab:ablation}
\setlength{\tabcolsep}{4pt}
\begin{tabular}{@{}lrrrrr@{}}
\toprule
& \multicolumn{2}{c}{AKI} & \multicolumn{2}{c}{Mortality} & Transfer QA\\
\cmidrule(lr){2-3}\cmidrule(lr){4-5}
Method / input & AUROC & AP & AUROC & AP & Acc. (\%)\\
\midrule
\textbf{\mname{}: full $V+R$} & \textbf{0.7440} & \textbf{0.4650} & \textbf{0.7350} & \textbf{0.2651} & \textbf{41.73}\\
\midrule
\multicolumn{6}{l}{\emph{Inference-time input interventions}}\\
Visual only: $V$ & 0.6250 & 0.2795 & 0.5872 & 0.1493 & 35.50\\
Residual only: $R$ & 0.5418 & 0.2216 & 0.5690 & 0.1332 & 29.95\\
\midrule
\multicolumn{6}{l}{\emph{Training ablations}}\\
Without zero initialization & 0.7303 & 0.4452 & 0.7205 & 0.2459 & 40.15\\
Without RMS scaling & 0.7357 & 0.4502 & 0.7250 & 0.2503 & 39.85\\
\bottomrule
\end{tabular}
\end{table}

\paragraph{Inference-time input interventions.}
We supply either the original visual tokens $V$ or the scaled numerical residual $R=s\odot\Delta$ at the visual positions. The trained adapter is unchanged. The residual-only input retains $s$ computed from $V$, as well as the original positional structure, so it still depends on visual information. Both interventions reduce prediction and QA scores relative to the full input. These results measure sensitivity to removing an input component; the altered inputs differ from those seen during training and do not isolate semantic use of chart content.

\paragraph{Zero initialization and RMS scaling.}
Removing zero initialization lowers AKI and mortality AP by 0.0198 and 0.0192; removing RMS scaling lowers them by 0.0148 and 0.0148. These are absolute AP differences. Both changes also reduce AUROC and transfer QA accuracy. This pattern supports starting adaptation from the native visual representation and expressing corrections relative to its feature scale for outcome prediction and QA.

\clearpage
\section{Category-Level Supervised QA}
\label{app:categories}
Table~\ref{tab:categories} gives the scores and differences underlying Figure~\ref{fig:categories}. All methods are trained on QA. We select the best of ChatTS, MLLM4TS, ITFormer, and OpenTSLM in each category, making the comparison at least as strong as any single baseline. \mname{}-Separate has higher mean accuracy than this reference in eight categories, ties in monitoring rules, and scores lower in ordering of changes and next-intervention prediction.
\begin{table}[!htbp]
\centering\footnotesize
\caption{\textbf{Supervised QA gains over the best baseline in each category at 4B.} All methods are trained on QA. The reference is the maximum of ChatTS, MLLM4TS, ITFormer, and OpenTSLM in each category. $\Delta$ is \mname{}-Separate minus this reference, in percentage points. Values include patient-bootstrap SD.}
\label{tab:categories}
\setlength{\tabcolsep}{3pt}
\begin{tabular}{lrrrr}
\toprule
Category & Best baseline (\%) & \mname{}-Joint (\%) & \mname{}-Separate (\%) & $\Delta$ (pp)\\
\midrule
ASR: Cross-variable retrieval & $41.50\pm 3.29$ & $52.00\pm 3.49$ & $53.00\pm 3.47$ & $11.50\pm 4.47$\\
CVR: Ordering of changes & $68.50\pm 3.25$ & $63.00\pm 3.42$ & $62.50\pm 3.42$ & $-6.00\pm 2.67$\\
IID: Next intervention & $83.50\pm 2.44$ & $78.00\pm 2.96$ & $80.50\pm 2.87$ & $-3.00\pm 2.27$\\
IR: Intervention response & $59.50\pm 3.40$ & $64.50\pm 3.39$ & $62.00\pm 3.40$ & $2.50\pm 4.30$\\
MA: Missingness counting & $59.00\pm 3.19$ & $54.00\pm 3.52$ & $63.00\pm 3.41$ & $4.00\pm 4.30$\\
MED: Monitoring rules & $75.50\pm 3.07$ & $76.00\pm 3.04$ & $75.50\pm 3.07$ & $0.00\pm 0.00$\\
NIF: Next-value interval & $47.00\pm 2.95$ & $66.00\pm 3.33$ & $66.50\pm 3.30$ & $19.50\pm 4.07$\\
TF: Threshold forecasting & $33.50\pm 2.93$ & $53.50\pm 3.60$ & $53.00\pm 3.56$ & $19.50\pm 4.37$\\
TG: Event localization & $66.50\pm 3.28$ & $77.00\pm 2.82$ & $77.00\pm 2.98$ & $10.50\pm 2.86$\\
TPR: Segmented trends & $58.00\pm 3.39$ & $70.50\pm 3.28$ & $71.50\pm 3.17$ & $13.50\pm 3.90$\\
TSS: Numerical summaries & $71.00\pm 2.91$ & $96.50\pm 1.33$ & $97.50\pm 1.10$ & $26.50\pm 3.07$\\
\bottomrule
\end{tabular}
\end{table}

\paragraph{Sensitivity to the numerical-summary category.}
Numerical summaries yield the highest accuracy and largest gain for \mname{}. Excluding this category leaves \mname{}-Separate at 66.45\%, compared with 59.25\% for the best baseline selected in each of the other ten categories. This 7.20 percentage-point difference therefore also holds against every individual baseline in the comparison.

\clearpage
\section{Majority-Answer and Random QA Baselines}
\label{app:qa_simple_baselines}
We evaluate two baselines that use no patient information on the same 2,200 test questions, with 200 questions per category. The \emph{majority-answer} baseline selects the most frequent answer letter (A, B, C, or D) among the 2,000 training questions in each category and always predicts that letter for the corresponding test category. Ties are resolved in alphabetical order. This measures answer-position preferences after option randomization. The \emph{uniform random} baseline selects each of the four options with probability $1/4$, giving an exact expected accuracy of 25\% in every category.

\begin{table}[!htbp]
\centering\small
\caption{\textbf{QA baselines without patient information.} Majority selects each category's most frequent training answer letter and reports test accuracy. Random is the exact expectation of uniform four-option guessing. All accuracies are percentages.}
\label{tab:qa_simple_baselines}
\setlength{\tabcolsep}{5pt}
\begin{tabular}{@{}lrcrr@{}}
\toprule
Category & Test $n$ & Majority letter & Majority (\%) & Random (\%)\\
\midrule
Event localization & 200 & A & 27.00 & 25.00\\
Cross-variable retrieval & 200 & A & 28.50 & 25.00\\
Segmented trends & 200 & A & 21.00 & 25.00\\
Missingness counting & 200 & A & 29.50 & 25.00\\
Numerical summaries & 200 & D & 25.50 & 25.00\\
Ordering of changes & 200 & C & 30.50 & 25.00\\
Intervention response & 200 & B & 22.50 & 25.00\\
Threshold forecasting & 200 & D & 25.00 & 25.00\\
Next-value interval & 200 & C & 22.50 & 25.00\\
Next intervention & 200 & A & 26.00 & 25.00\\
Monitoring rules & 200 & A & 23.00 & 25.00\\
\midrule
Macro-average & 2200 & - & 25.55 & 25.00\\
\bottomrule
\end{tabular}
\end{table}

The majority-answer baseline has macro-average accuracy of 25.55\%, ranging from 21.00\% to 30.50\% across categories. Uniform random guessing has expected macro-average accuracy of 25.00\%. These references complement the model comparisons in Table~\ref{tab:categories}; the majority letter is chosen using training answers only.

\clearpage

\section{Specialized Clinical Prediction References}
\label{app:references}
The specialized prediction systems in Table~\ref{tab:specialists} contextualize the clinical endpoint scores. LSTM, GRU, ConCare, VITA, and ColaCare exceed the 9B \mname{}-Joint interface on both endpoints. Their task-specific output heads produce endpoint scores; answering new temporal questions requires an additional language component. \mname{} retains the pretrained language head, allowing the prediction-trained interface to be evaluated on QA. At 9B, \mname{}-Joint also has higher mean scores than the specialized Transformer reference on all four prediction metrics.
\begin{table}[!htbp]
\centering\small
\caption{\textbf{Specialized clinical prediction references.} Reported AUROC/AP $\pm$ SD. These systems have no QA evaluation in this study.}
\label{tab:specialists}
\setlength{\tabcolsep}{4pt}
\begin{tabular}{lrrrr}
\toprule
& \multicolumn{2}{c}{AKI} & \multicolumn{2}{c}{Mortality}\\
Method & AUROC & AP & AUROC & AP\\
\midrule
LSTM & $0.7793\pm 0.0178$ & $0.5244\pm 0.0354$ & $0.7780\pm 0.0212$ & $0.3357\pm 0.0473$\\
GRU & $0.7761\pm 0.0183$ & $0.5170\pm 0.0341$ & $0.7728\pm 0.0233$ & $0.3248\pm 0.0474$\\
Transformer & $0.7378\pm 0.0183$ & $0.4400\pm 0.0340$ & $0.7288\pm 0.0256$ & $0.2997\pm 0.0465$\\
ConCare & $0.7946\pm 0.0175$ & $0.5367\pm 0.0331$ & $0.7881\pm 0.0218$ & $0.3505\pm 0.0484$\\
VITA & $0.8006\pm 0.0170$ & $0.5429\pm 0.0354$ & $0.7969\pm 0.0211$ & $0.3569\pm 0.0485$\\
ColaCare & $0.8074\pm 0.0170$ & $0.5506\pm 0.0328$ & $0.8020\pm 0.0210$ & $0.3615\pm 0.0491$\\
\bottomrule
\end{tabular}
\end{table}




\section{Prediction Probability Calibration}
\label{app:calibration}
We evaluate the model's positive-event probabilities directly, without post-hoc recalibration. For outcome $y_i\in\{0,1\}$ and predicted probability $p_i$, the Brier score is
\begin{equation}
 \operatorname{Brier}=\frac{1}{n}\sum_{i=1}^{n}(p_i-y_i)^2.
\end{equation}
ECE partitions $[0,1]$ into ten equal-width bins $B_b$ and compares each bin's mean probability with its event frequency:
\begin{equation}
 \operatorname{ECE}=\sum_{b=1}^{10}\frac{|B_b|}{n}
 \left|\frac{1}{|B_b|}\sum_{i\in B_b}p_i-
       \frac{1}{|B_b|}\sum_{i\in B_b}y_i\right|.
\end{equation}
Empty bins contribute zero. This measures calibration of the predicted event probability. Brier score evaluates overall probability error and reflects both calibration and discrimination. Lower values are better for both metrics. Standard deviations use 2,000 patient-level bootstrap resamples with fixed bin boundaries.

\paragraph{Interpreting native probabilities.}
All Qwen methods use the same label convention: $1$ denotes the event and $0$ its absence, and $p_i$ is the normalized probability of label $1$. For native models, these answer probabilities have not been fitted to clinical outcomes or event prevalence. Answer preferences can therefore produce extreme probabilities and large calibration errors \citep{contextual_calibration}. The large 9B native-chart mortality errors characterize this zero-shot use of label probabilities; they are distinct from the weak patient ranking measured by AUROC. The main prediction comparison also includes clinically trained Chart LoRA and Text LoRA baselines.

\begin{table}[!htbp]
\centering\small
\caption{\textbf{Numerical adaptation improves prediction probability quality.} Probabilities are evaluated without recalibration; lower is better. Bold marks the minimum for each model size, outcome, and metric. Table~\ref{tab:calibration_full} reports both unadapted input baselines, including patient-bootstrap SD.}
\label{tab:calibration}
\setlength{\tabcolsep}{3pt}
\begin{tabular}{lrrrrrr}
\toprule
& \multicolumn{3}{c}{AKI} & \multicolumn{3}{c}{Mortality}\\
\cmidrule(lr){2-4}\cmidrule(lr){5-7}
Scale & Unadapted plots & \mname{}-Joint & \mname{}-Separate & Unadapted plots & \mname{}-Joint & \mname{}-Separate\\
\midrule
\multicolumn{7}{l}{\emph{ECE $\downarrow$}}\\
2B & 0.1965 & 0.0468 & \textbf{0.0444} & 0.4962 & \textbf{0.0328} & 0.0338\\
4B & \textbf{0.0378} & 0.0437 & 0.0614 & 0.3860 & 0.0383 & \textbf{0.0373}\\
9B & 0.2418 & 0.0820 & \textbf{0.0511} & 0.7261 & 0.0666 & \textbf{0.0284}\\
\midrule
\multicolumn{7}{l}{\emph{Brier score $\downarrow$}}\\
2B & 0.2026 & 0.1461 & \textbf{0.1454} & 0.3482 & \textbf{0.0937} & 0.0939\\
4B & 0.1590 & \textbf{0.1395} & 0.1474 & 0.2537 & \textbf{0.0911} & 0.0935\\
9B & 0.2364 & 0.1437 & \textbf{0.1412} & 0.6257 & 0.0922 & \textbf{0.0915}\\
\bottomrule
\end{tabular}
\end{table}

\begin{table}[!htbp]
\centering\footnotesize
\caption{\textbf{Calibration of unadapted models with text and chart inputs.} Values are mean $\pm$ patient-bootstrap SD. ECE uses ten equal-width probability bins. Bold marks the smallest mean within each scale and metric.}
\label{tab:calibration_full}
\setlength{\tabcolsep}{4pt}
\begin{tabular}{llrrrr}
\toprule
& & \multicolumn{2}{c}{AKI} & \multicolumn{2}{c}{Mortality}\\
Scale & Method & ECE $\downarrow$ & Brier $\downarrow$ & ECE $\downarrow$ & Brier $\downarrow$\\
\midrule
2B & Unadapted text & 0.3462 $\pm 0.0133$ & 0.2895 $\pm 0.0021$ & 0.5665 $\pm 0.0096$ & 0.4214 $\pm 0.0044$\\
 & Unadapted plots & \textbf{0.1965} $\pm 0.0130$ & \textbf{0.2026} $\pm 0.0030$ & \textbf{0.4962} $\pm 0.0097$ & \textbf{0.3482} $\pm 0.0035$\\
\midrule
4B & Unadapted text & 0.0861 $\pm 0.0124$ & 0.1780 $\pm 0.0059$ & \textbf{0.1300} $\pm 0.0095$ & \textbf{0.1127} $\pm 0.0048$\\
 & Unadapted plots & \textbf{0.0378} $\pm 0.0111$ & \textbf{0.1590} $\pm 0.0080$ & 0.3860 $\pm 0.0100$ & 0.2537 $\pm 0.0040$\\
\midrule
9B & Unadapted text & 0.4843 $\pm 0.0145$ & 0.2990 $\pm 0.0033$ & \textbf{0.4502} $\pm 0.0092$ & \textbf{0.4586} $\pm 0.0088$\\
 & Unadapted plots & \textbf{0.2418} $\pm 0.0136$ & \textbf{0.2364} $\pm 0.0046$ & 0.7261 $\pm 0.0095$ & 0.6257 $\pm 0.0071$\\
\bottomrule
\end{tabular}
\end{table}

Table~\ref{tab:calibration_full} reports results for the unadapted models with text and chart inputs. Relative to the unadapted chart baseline, both \mname{} encodings lower Brier score at every model size and for both outcomes (Table~\ref{tab:calibration}). ECE also decreases except for 4B AKI, where the unadapted chart model already has low ECE. Reporting both metrics distinguishes improvements in probability error from uniform gains in calibration.

\end{document}